\documentclass[10pt]{article}

\usepackage{amsmath}
\usepackage{amssymb}

\usepackage{graphicx}
\usepackage{hyperref}
\usepackage{cite}

\usepackage{color}

\usepackage{setspace}

\usepackage[labelfont=bf,labelsep=period,justification=raggedright]{caption}

\makeatletter
\renewcommand{\@biblabel}[1]{\quad#1.}
\makeatother

\date{}

\usepackage{threeparttable}

\begin{document}

\begin{flushleft}
{\Large
\textbf{A practical approach to language complexity: A Wikipedia case study}
}
\\
Taha Yasseri$^{1,\ast}$,
Andr\'{a}s Kornai$^{2}$,
J\'{a}nos Kert\'{e}sz$^{1,3}$
\\
\bf{1} Department of Theoretical Physics, Budapest University of Technology and Economics, Budapest, Hungary
\\
\bf{2} Computer and Automation Research Institute, Hungarian Academy of Sciences, Budapest, Hungary
\\
\bf{3} Center for Network Science, Central European University, Budapest, Hungary
\\
$\ast$ E-mail: yasseri@phy.bme.hu
\end{flushleft}

\section*{Abstract}
 In this paper we present statistical analysis of English texts from
 Wikipedia.  We try to address the issue of language complexity empirically by
 comparing the simple English Wikipedia (Simple) to comparable samples of the
 main English Wikipedia (Main).  Simple is supposed to use a more simplified
 language with a limited vocabulary, and editors are explicitly requested to
 follow this guideline, yet in practice the vocabulary richness of both
 samples are at the same level. Detailed analysis of longer units (n-grams of
 words and part of speech tags) shows that the language of Simple is less
 complex than that of Main primarily due to the use of shorter sentences, as
 opposed to drastically simplified syntax or vocabulary.  Comparing the two
 language varieties by the Gunning readability index supports this conclusion.
 We also report on the topical dependence of language complexity, e.g.  that
 the language is more advanced in conceptual articles compared to person-based
 (biographical) and object-based articles. Finally, we investigate the
 relation between conflict and language complexity by analyzing the content of
 the talk pages associated to controversial and peacefully developing
 articles, concluding that controversy has the effect of reducing language
 complexity.

\section*{Introduction}
Readability is one of the central issues of language complexity and applied
linguistics in general \cite{orlow2003}.  Despite the long history of
investigations on readability measurement, and significant effort to introduce
computational criteria to model and evaluate the complexity of text in the
sense of readability, a conclusive and fully representative scheme is still
missing \cite{klare1974,kanungo2009,karamakar2010}.  In recent years the large
amount of machine readable user generated text available on the web has
offered new possibilities to address many classic questions of
psycholinguistics. Recent studies, based on text-mining of blogs
\cite{lambiotte2007}, web pages \cite{serrano2009}, online forums
\cite{altmann2009,altmann2011}, etc, have advanced our understanding of
natural languages considerably.

Among all the potential online corpora, Wikipedia, a multilingual online
encyclopedia \cite{wiki}, which is written
collaboratively by volunteers around the world, has a special position.  Since
Wikipedia content is produced collaboratively, it is a uniquely unbiased sample. As
Wikipedias exist in many languages, we can carry out a wide range of cross-linguistic
studies. Moreover, the broad studies on social aspects of Wikipedia and its communities
of users \cite{voss2005,ortega2007,halavais2008,Javanmardi2010b,laniado2011b,massa2011,kimmons2011,yasseri-circ,yasseri2012} makes it
possible to develop  sociolinguistic descriptions for the linguistic observations.

One of the particularly interesting editions of Wikipedia is the {\it
  Simple English Wikipedia} \cite{wiki-simple}
(Simple). Simple aims at providing an encyclopedia for people with only basic
knowledge of English, in particular children, adults with learning
difficulties, and people learning English as a second language. See
  Table~\ref{tab:april} comparing the articles for `April' in Simple and Main. In this work,
we reconsider the issue of language complexity based on the statistical
analysis of a corpus extracted from Simple. We compare basic measures of
readability across Simple and the standard English Wikipedia (Main) \cite{wiki-main} to understand how
simple is Simple in comparison. Since there are no supervising editors
involved in the process of writing Wikipedia articles, both Simple and Main are
uncorrected (natural) output of the human language generation ability. The
text of Wikipedias is emerging from contributions of a large number of independent
editors, therefore all different types of personalization and bias are
eliminated, making it possible to address the fundamental concepts regardless
of marginal phenomena.

Readability studies on different corpora have a long history; see \cite{baumann2005}
for a summary. In a recent study \cite{roberts1994}, readability
of articles published in the {\it Annals of Internal Medicine} before and
after the reviewing process is investigated, and a slight improvement in
readability upon the review process is reported.  Wikipedia is widely used to extract concepts,
relations, facts and descriptions by applying natural language processing
techniques \cite{medelyan2009}.
In \cite{gabrilovich2007,zesch2008,wang2008,gabrilovich2009} different authors have tried to
extract semantic knowledge from Wikipedia aiming at measuring semantic relatedness,
lexical analysis and text classification. Wikipedia is used to
establish topical indexing methods in \cite{medelyan2008}. Tan and Fuchun performed query
segmentation by combining generative language models and Wikipedia
information \cite{tan2008}. In a novel approach, Tyers and Pienaarused used Wikipedia to extract bilingual word pairs from
interlingual hyperlinks connecting articles from different language editions \cite{tyers2008}.
And more practically, Sharoff and Hartley have been seeking for ``suitable texts for language
learners'', developing a new complexity measure, based on both lexical and
grammatical features \cite{sharoff2008}.
Comparisons between Simple and Main for the selected set of articles show
that in most cases Simple has less complexity, but there exist exceptional
articles, which are more readable in Main than in Simple.  In a complementary
study \cite{besten2008}, Simple is examined by measuring the Flesch reading score
\cite{flesch1979}. They found that Simple is not simple enough compared to
other English texts, but there is a positive trend for the whole Wikipedia to become
more readable as time goes by, and that the tagging of those articles that
need more simplifications by editors is crucial for this achievement.
In a new class of applications \cite{napoles2010,yatskar2010,coster2011}, Simple is used to establish automated text simplification algorithms.

\begin{table}[ht!]
\caption{
\bf{The articles on {\it April} in Main English and Simple English Wikipedias.}}
\begin{tabular}{p{.4 \textwidth}p{.4 \textwidth}}
Main & Simple \\
April is the fourth month of the year in the Julian and Gregorian calendars, and one of four months with a length of 30 days.
The traditional etymology is from the Latin aperire, ``to open,"
in allusion to its being the season when trees and flowers begin to ``open". & April is the fourth month of the year.
 It has 30 days. The name April comes
from that Latin word aperire which means ``to open". This probably refers to growing plants in spring.\\
\end{tabular}
\label{tab:april}
\end{table}


\section*{Methods}

We built our own corpora from the dumps \cite{wiki-dumps} of Simple and Main
Wikipedias released at the end of 2010 using the WikiExtractor developed at
the University of Pisa Multimedia Lab (see Text S2 in the Supporting
Information for the availability of this and other software packages and
corpora used in this work). The Simple corpus covers the whole text of Simple
Wikipedia articles (no talk pages, categories and templates). For the Main
English Wikipedia, first we made a big single text including all articles, and
then created a corpus comparable to Simple by randomly selecting texts having
the same sizes as the Simple articles. In both samples HTML entities were
converted to characters, MediaWiki tags and commands were discarded, but the
anchor texts were kept.

\begin{table}[ht!]
\caption{
\bf{Vocabulary richness in Main and Simple}}
\begin{tabular}{ccccc}
Cond & SR & $C_M$ & $C_S$ & $C_M/C_S$\\
CB & 1.0002  & 0.8226 & 0.8167 & 1.0072\\
CN & 0.9997  & 0.7782 & 0.7739 & 1.0055\\
WB & 1.0000  & 0.8218 & 0.8167 & 1.0061\\
WN & 1.0000  & 0.7774 & 0.7739 & 1.0045\\
CBP & 1.0002 & 0.8061 & 0.8013 & 1.0059\\
CNP & 0.9997 & 0.7568 & 0.7542 & 1.0034\\
WBP & 1.0000 & 0.8052 & 0.8013 & 1.0049\\
WNP & 1.0000 & 0.7563 & 0.7543 & 1.0028\\
\end{tabular}
\begin{flushleft} 
For the definition of conditions (character- or word-balanced, with or without
puctuation, with or without Porter stemming) see the Methods section. SR is
size ratio (number of characters in C conditions, number of words in W
conditions) for comparable Main and Simple corpora. $C_M$ and $C_S$ are
Herdan's $C$ for Main and Simple. As the last column shows, the vocabulary
richness of comparable Simle and Main corpora differs at most by 0.72\%
depending on condition.
\end{flushleft}
\label{tab:stat}
\end{table}

Simple uses significantly shorter words (4.68 characters/word) than Main (5.01
characters/word). We can define `same size' by equal number of characters (see
Condition CB in Table~\ref{tab:stat}), or by equal number of words (Condition
WB). Since sentence lengths are also quite different (Simple has 17.0
words/sentence on average, Main has 25.2), the standard practice of
computational linguistics of counting punctuation marks as full word tokens
may also be seen as problematic. Therefore, we created two further conditions,
CN (character-balanced but no punctuation) and WN (word-balanced no
punctuation). In both conditions, we used the standard (Koehn, see Text S2) tokenizer to
find the words, but in the N conditions we removed the punctuation chars {\tt
  ,.?();"!:}. Another potential issue concerns stemming, whether we consider
the tokens {\it amazing, amazed, amazes} as belonging to the same or different
types. To see whether this makes any difference, we also created conditions
CBP, WBP, CNP, and WNP by stemming both Simple and Main using the standard
Porter stemmer \cite{PorterStemmer}. Table~\ref{tab:stat} compares for Simple
and Main a classic measure of vocabulary richness, Herdan's $C$, defined as
log(\#types)/log(\#tokens), under these conditions. 

For word and part of speech (POS) n-gram statistics not all these conditions
make sense, since automatic POS taggers crucially rely on information in the
affixes that would be destroyed by stemming, and for the automatic detection
of sentence boundaries punctuation is required \cite{Mikheev:2002a}. We
therefore used word-balanced samples with punctuation kept in place (condition
WB) but distinguished different conditions of POS tagging for the following
reason. Wikipedia, and encyclopedias in general, use an extraordinary amount
of proper names (three times as much as ordinary English as measured e.g. on
the Brown Corpus), many of which are multiword {\it named entities}. An
ordinary POS tagger may not recognize that Long Island is a single named
entity and could tag it as JJ NN (adjective noun) rather than as NNP NNP
(proper name phrase).  Therefore, we supplemented the original POS tagging
(Condition O) by a named entity recognition (NER) system and rerun the POS
tagging in light of the NER output (Condition N).  If adjacent NNP-tagged
elements are counted as a single NE phrase, we obtain the SO (shortened
original) and SN (shortened NER-based) versions. Since neither word-based nor
POS-based n-grams are very meaningful if they span sentence boundaries, we
also created `postprocessed' versions, where for odd n those n-grams where the
boundary was in the middle were omitted, and the words/tags falling on the
shorter side were uniformly replaced by the boundary marker both for odd and
even n.

To measure text readability, we limited ourselves to the ``Gunning fog index''
$F$, \cite{gunning1952,gunning1969} which is one of the simplest and most
reliable among all different recent and classic measures (see
\cite{kincaid1975,collins2004,dubay2007}). $F$ is calculated as

$$ F= 0.4(\frac{\textrm{\#words}}{\textrm{\#sentences}}+100\frac{
  \textrm{\#complex words}}{\textrm{\#words}})$$

\noindent
where words are considered complex if they have three or more syllables. A
simple interpretation of $F$ is the number of years of formal education needed
to understand the text.

\section*{Results and Discussion}

We present our results in three parts. First we report on overall comparison
of Main and Simple at different levels of word and n-gram statistics in
addition to readability analysis. Next we narrow down the analysis further to
compare selected articles and categories of articles, and examine the dependence
of language complexity on the text topic. Finally, we explore the relation
between controversy and language complexity by considering the case of
editorial wars and related discussion pages in Wikipedia.

\subsection*{Overall comparison}
\subsubsection*{Readability}
In Table~\ref{tab:fog}, the Gunning fog index calculated for 6 different English corpora is reported.
Remarkably, the fog index of Simple is higher than that of Dickens, whose
writing style is sophisticated but doesn't rely on the use of longer latinate
words which are hard to avoid in an encyclopedia. The British National Corpus,
which is a reasonable approximation to what we would want to think of as
`English in general' is a third of the way between Simple and Main, demonstrating the
accomplishments of Simple editors, who pushed Simple half as much below average
complexity as the encyclopedia genre pushes Main above it.

\begin{table}[ht!]
\caption{
\bf{Readability of different English corpora}}
\begin{tabular}{cccc}
 Corpus         &  $F$            & Corpus        & $F$  \\
Dickens        &$8.6 \pm0.1$ & Simple    &$10.8 \pm0.2$ \\
SJM        &$10.3 \pm0.1$      & BNC        &$12.1 \pm0.5$ \\
WSJ         &$10.8 \pm0.2$     & Main        &$15.8 \pm0.4$ \\
\end{tabular}
\begin{flushleft}
Gunning fog index for 6 different corpora of WSJ: {\it Wall Street Journal$^\bullet$\tnote{wsj}}, {\it Charles Dickens'} books,
SJM: {\it San Jose Mercury News$^*$\tnote{sjm}}, BNC: {\it British National Corpus$^\dagger$\tnote{sjm}},
Simple, and Main.
\begin{tablenotes}
\item[wsj] $^\bullet$http://www.wsj.com
\item[sjm] $^*$http://www.mercurynews.com
\item[sjm] $^\dagger$http://www.natcorp.ox.ac.uk
\end{tablenotes}
\end{flushleft}
\label{tab:fog}
\end{table}

\subsubsection*{Word statistics}

Vocabulary richness is compared for Simple and Main in Table~\ref{tab:stat}
using Herdan's $C$, a measure that is remarkably stable across sample sizes:
for example using only 95\% of the word-balanced (Condition WB) samples we
would obtain $C$ values that differ from the ones reported here by less than
0.066\% and 0.044\%. For technical reasons we could not balance the samples
perfectly (there is no sense in cutting in the middle of a line, let alone the
middle of a word), but the size ratios (column SR in Table~\ref{tab:stat})
were kept within 0.03\%, two orders of magnitude less discrepancy than the 5\%
we used above, making the error introduced by less than perfect balancing
negligible.

The precise choice of condition has a significant impact on $C$, ranging from
a low of 0.754 (character-balanced, no punctuation, Porter stemming) to a high
of 0.8226 (character-balanced, punctuation included, no stemming), but
practically no effect on the $C_M/C_S$ ratio, which is between 0.28\% and
0.72\% for all conditions reported here. In other words, we observe the same
vocabulary richness in balanced samples of Simple and Main quite independent
of the specific processing and balancing steps taken.  We also experimented
with several other tokenizers and stemmers, as well as inclusion or exclusion
of numerals or words with foreign (not ISO-8859-1) characters, but the precise
choice of condition made little difference in that the discrepancy between
$C_M$ and $C_S$ always stayed less than 1\% ($-0.27\%$ to $+0.72\%$).  The
only condition where a more significant difference of 3.4\% could be observed
was when Simple was directly paired with Main by selecting, wherever possible,
the corresponding Main version of every Simple article. 

\begin{figure*}[ht!]
\includegraphics[width=0.95\textwidth]{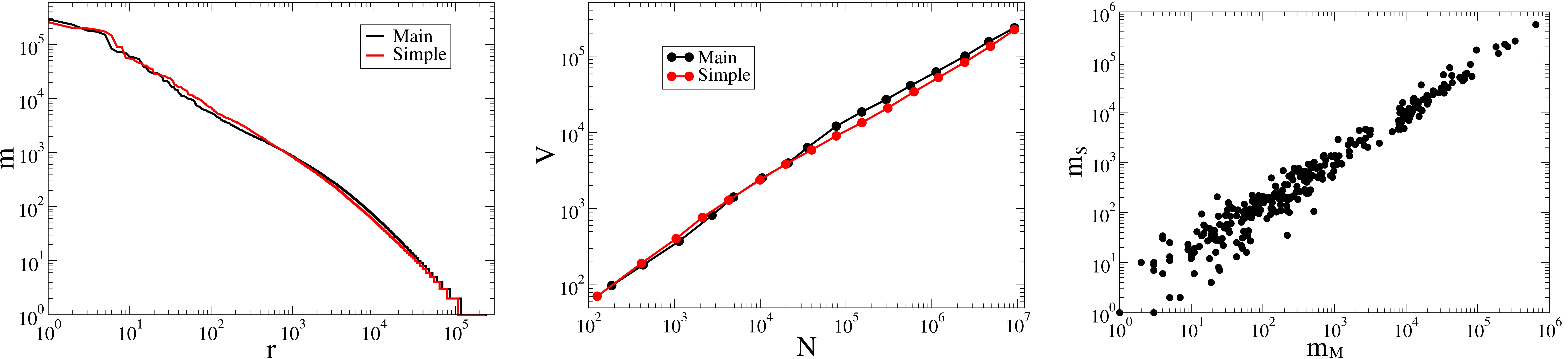}
\caption{{\bf Word-level statistical analysis of Main and Simple.} Condition WB, as explained the Methods section. {\it left:}
  Zipf's law for the Main (black) and Simple (red) samples.  {\it middle:}
  Heaps' law (same colors). The exponents are $0.72\pm0.01$ (Main) and
  $0.69\pm0.01$ (Simple).  {\it right:} Comparing token frequencies in the two
  samples for 300 randomly selected words (``S'' and ``M'' stand for Simple
  and Main respectively), the correlation coefficient is C=0.985.  All three
  diagrams show that the two samples have statistically almost the same
  vocabulary richness.}\label{fig:1gram}
\end{figure*}

As discussed in \cite{tweedie1998}, one cannot reasonably expect the same
result to hold for other traditional measures of vocabulary richness such as
type-token ratio, since these are not independent of sample size
asymptotically \cite{kornai2002}. However, Herdan's Law (also known as Heaps'
Law, \cite{Herdan:1964,heaps1978}), which states that the number of different
types $V$ scales with the number of tokens $N$ as $ V \sim N^C$, is known to
be asymptotically true for any distribution following Zipf's law
\cite{zipf1935}, see \cite{Kornai:1999,baeza2000,vLvdW}.  In
Fig.~\ref{fig:1gram} (left and middle panels) our study of both laws in Condition WB, are
illustrated. 

Since all these results demonstrate the similarity of the Simple and Main
samples in the sense of unigram vocabulary richness, a conclusion that is
quite contrary to the Simple Wikipedia stylistic guidelines
\cite{wiki-how-to-write-simple}, we performed some additional tests.  First,
we selected 300 words randomly and compared the number of their appearance in
both samples (right panel of Fig.~\ref{fig:1gram}). Next, we considered the
word entropy of Simple and Main, obtaining 10.2 and 10.6 bits
respectively. Again, the exact numbers depend on the details of preprocessing,
but the difference is in the 2.9\% to 3.9\% range in favor of Main in every
condition, while the dependence on condition is in the 1.8\% to 2.8\% range.
Though 0.4 bits are above the noise level, the numbers should be compared to
the word entropy of mixed text, 9.8 bits, as measured on the Brown Corpus, and
of spoken conversation, 7.8 bits, as measured on the Switchboard Corpus. When
a switch in genre can bring over 30\% decrease in word entropy, a 3\%
difference pales in comparison. Altogether, both Simple and Main are close in
word entropy to high quality newspaper prose such as the Wall Street Journal,
10.3 bits, and the San Jose Mercury News, 11.1 bits.

\subsubsection*{Word n-gram statistics}\label{sec:ngram}

\begin{figure*}[ht!]
\includegraphics[width=0.95\textwidth]{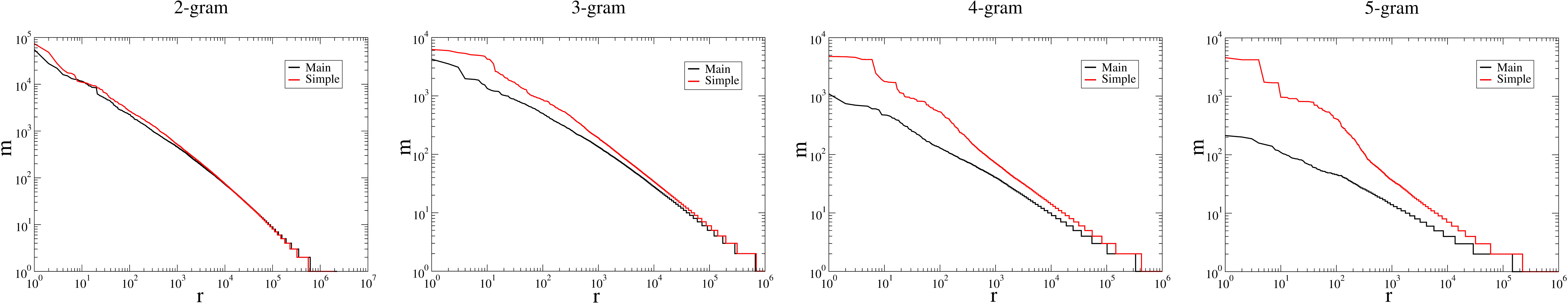}
\caption{{\bf N-gram statistical analysis of Main and Simple.} Condition WB, as explained the Methods section. Number of appearances of n-grams in Main (black) and Simple (red) for $n = \textrm{$2$--$5$}$ from left to right.
By increasing $n$, the difference between
two samples becomes more significant. In Simple there are more of the
frequently appearing n-grams than in Main.}\label{fig:ngrams}
\end{figure*}

One effect not measured by the standard unigram techniques is the contribution
of lexemes composed of more than one word, including idiomatic expressions
like `take somebody to task' and collocations like `heavy drinker'. The Simple
Wikipedia guidelines \cite{wiki-how-to-write-simple} explicitly warn against
the use of idioms: `Do not use idioms (one or more words that together mean
something other than what they say)'. One could assume that Simple editors
rely more on such multiword patterns, and the n-gram analysis presented here
supports this. In Fig.~\ref{fig:ngrams} made under condition WB, the token frequencies of n-grams are
shown in a Zipf-style plot as a function of their rank.  Both the unigram
statistics discussed in the previous section and the 2-gram statistics
presented here are nearly identical for Simple and Main, but 3-grams and
higher n-grams begin to show some discrepancy between them. In reality, a
sample of this small size (below $10^7$ words) is too small to represent
higher n-grams well, as is clear from manual inspection of the top 5-grams of
Simple.  

Ignoring 5-grams composed of Chinese characters (which are mapped into the
same string by the tokenizer), the top four entries, with over 4200
occurrences, all come from the string {\tt . It is found in the region}.  In
fact, by grepping on high frequency n-grams such as {\it is a commune of} we
find over six thousand entries in Simple such as the following: {\it Alairac
  is a commune of 1,034 people (1999). It is located in the region
  Languedoc-Roussillon in the Aude department in the south of France.}  Since
most of these entries came from only a handful of editors, we can be
reasonably certain that they were generated from geographic databases
(gazetteers) using a simple `American Chinese Menu' substitution tool
\cite{Sproat:2010}, perhaps implemented as Wikipedia robots.

Since an estimated 12.3\% of the articles in Simple fit these patterns, it is
no surprise that they contribute somewhat to the apparent n-gram simplicity of
Simple.  Indeed, the entropy differential between Main and Simple, which is
0.39 bits absolute (1.7\% relative) for 5-grams, decreases to 0.28 bits (1.2\%
relative) if these articles are removed from Simple and the Main sample is
decreased to match. (By word count the robot-generated material is less than
2\% of Simple, so the adjustment has little impact.)  Since higher n-grams are
seriously undersampled (generally, $10^9$ words `gigaword corpora' are
considered necessary for word trigrams, while our entire samples are below
$10^7$ words) we cannot pursue the matter of multiword patterns further, but
note that the boundary between the machine-generated and the manually written
is increasingly blurred.

Consider {\it Joyeuse is a commune in the French department of Ard\`{e}che in
  the region of Rh\^{o}ne-Alpes. It is the seat of the canton of Joyeuse}, an
article that clearly started its history by semi-automatic or fully automatic
generation. By now (August 2012) the article is twice as long (either by
manual writing or semi-automatic import from the main English wikipedia), and
its content is clearly beyond what any gazetteer would list.  With high
quality robotic generation, editors will simply not know, or care, whether
they are working on a page that originally comes from a robot.  Therefore, in
what follows we consider Simple in its entirety, especially as the part of
speech (POS) statistics that we now turn to are not particularly impacted by
robotic generation.

\subsubsection*{Part of speech statistics}

\begin{figure*}[ht!]
\includegraphics[width=0.75\textwidth]{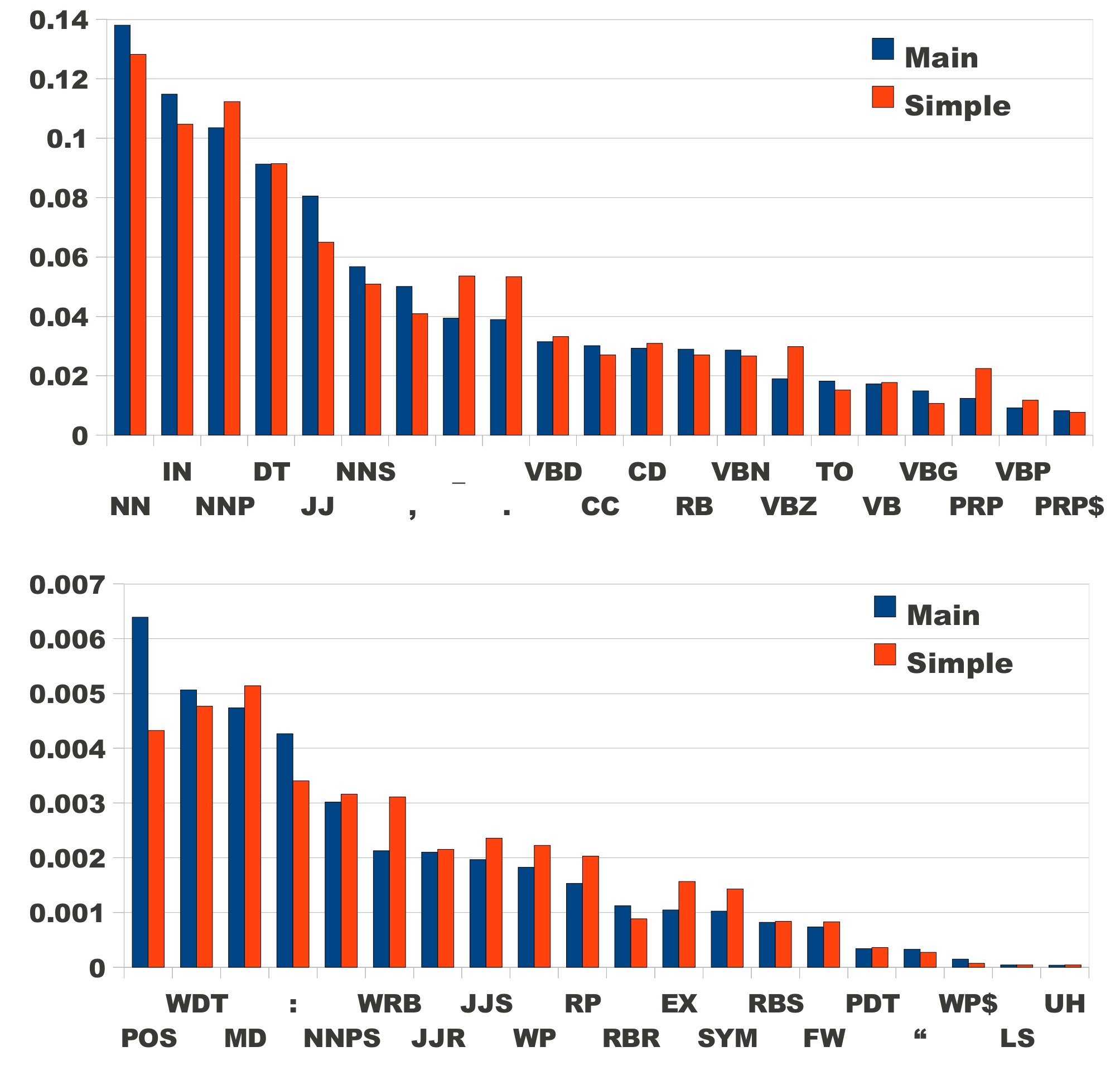}
\caption{{\bf Part of Speech statistics of Main English and Simple English Wikipedias.} Condition O, as explained the Methods section. The legends are defined as
NN: Noun, singular or mass; IN: Preposition or subordinating conjunction; NNP: Proper noun, singular; 
DT: Determiner; JJ: Adjective; NNS: Noun, plural; VBD: Verb, past tense; CC: Coordinating conjunction;
CD: Cardinal number; RB: Adverb; VBN: Verb, past participle;
VBZ: Verb, 3rd person singular present; TO: to; VB: Verb, base form; VBG: Verb, gerund or present participle;
PRP: Personal pronoun; VBP: Verb, non-3rd person singular present; PRP\$: Possessive pronoun;
POS: Possessive ending; WDT: Wh-determiner; MD: Modal; NNPS: Proper noun, plural; WRB: Wh-adverb; JJR: Adjective, comparative;
JJS: Adjective, superlative; WP: Wh-pronoun; RP: Particle; RBR: Adverb, comparative; EX: Existential there; SYM: Symbol;
RBS: Adverb, superlative; FW: Foreign word; PDT: Predeterminer;
WP\$: Possessive wh-pronoun; LS: List item marker; UH: Interjection;}\label{fig:pos}
\end{figure*}

Figure~\ref{fig:pos} shows the distribution of the part of speech (POS) tags
in Main and Simple for Condition O (word balanced, punctuation and possessive
{\it 's} counted as separate words, as standard with the the Penn Treebank POS
set \cite{pos-set}.) It is evident from comparing the first and second columns
that the encyclopedia genre is particularly heavy on Named Entities (proper
nouns or phrases designating specific places, people, and organizations
\cite{chinchor98}). Since multiword entities like {\it Long Island, Benjamin
  Franklin, National Academy of Sciences} are quite common, we also
preprocessed the data using the {\tt HunNER} Named Entity Recognizer
\cite{hunner}, and performed the part of speech tagging afterwards (condition
N). When adjacent NNP words are counted as one, we obtained the SO and SN
conditions. This obviously affects not just the NNP counts, but also the
higher n-grams that contain NNP.

Again, the similarity of Simple and Main is quite striking: the cosine
similarity measure of these distributions is between 0.989 (Condition O) and
0.991 (Condition SO), corresponding to an angle of 7.7 to 8.6 degrees.  To put
these numbers in perspective, note that the similarity between Main and the
Brown Corpus is 0.901 (25.8 degrees), and between Main and Switchboard 0.671
(47.8 degrees). For POS n-grams, it makes sense to omit n-grams with a
sentence boundary at the center. For the POS unigram models this means that we
do not count the notably different sentence lengths twice, a step that would
bring cosine similarity between Simple and Main to 0.992 (Condition SO) or 0.993
(Condition N), corresponding to an angle of 6.8 to 7.1 degrees. Either way, 
the angle between Simple and Main is remarkably acute. 

While Figure~3 shows some slight stylistic variation, e.g. that Simple uses
twice as many personal pronouns ({\it he, she, it, ...}) as Main, it is hard
to reach any overarching generalizations about these, both because most of the
differences are statistically insignificant, and because they point in
different directions. One may be tempted to consider the use of pronouns to be
an indicator of simpler, more direct, and more personal language, but by the
same token one would have to consider the use of wh-adverbs ({\it how however
  whence whenever where whereby wherever wherein whereof why ...})  to be a
hallmark of more sophisticated, more logical, and more impersonal style, yet
it is Simple that has 50\% more of these.

\begin{figure*}[ht!]
\includegraphics[width=0.75\textwidth]{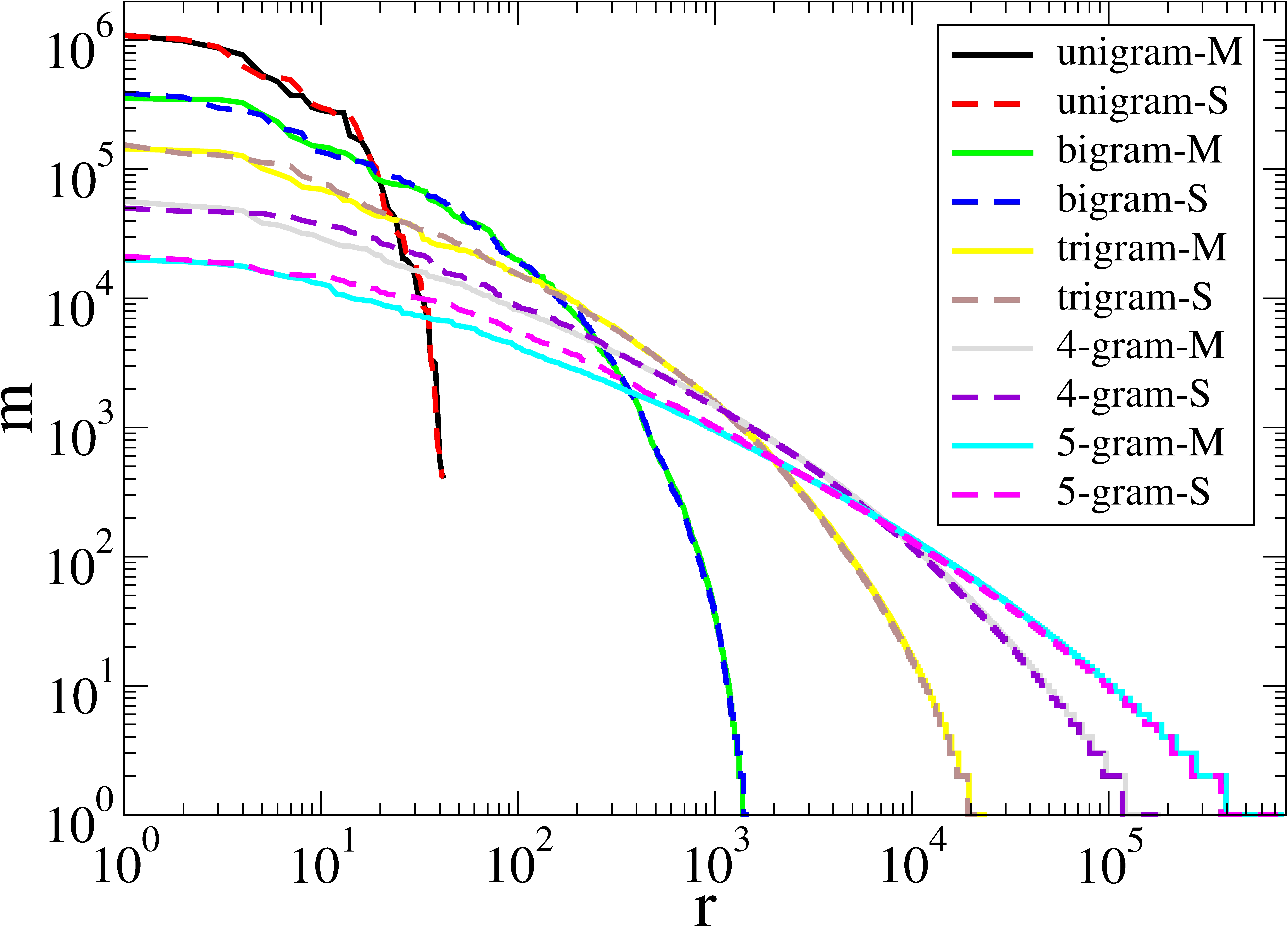}
\caption{{\bf POS-N-gram statistical analysis of
Main and Simple} Number of appearances of POS n-grams in Main and Simple for $n = \textrm{$1$--$5$}$ under condition N.}\label{fig:pos-all}
\end{figure*}

Figure~\ref{fig:pos-all} shows that the POS n-gram Zipf plots for $n=1,
\ldots, 5$ are practically indistinguishable across Simple and Main under
Condition N. (We are publishing this figure as it is the worst -- under the
other conditions, the match is even better.) In terms of cosine similarity,
the same tendencies that we established for unigram data remain true for
bigram or higher POS n-grams: the Switchboard data is quite far from both
Simple and Main, the Brown Corpus is closer, and the WSJ is closest. However,
Simple and Main are noticeably closer to one another than either of them is to
WSJ, as is evident from the Table~\ref{tab:angle}, which gives the angle, in decimal
degrees, between Simple and Main (column SM), Main and WSJ (column MW), and
Simple and WSJ (column SW) based on POS n-grams for $n= 2,\ldots,5$, under
condition SN, with postprocessing of n-grams spanning sentence boundaries. We
chose this condition because we believe it to be the least noisy, but we
emphasize that the same relations are observed for all other conditions, with
or without sentence boundary postprocessing, with or without removal of
machine-generated entries from Simple, with or without readjusting the Main
corpus to reflect this change (all 32 combinations were investigated).
The data leave no doubt that the WSJ is closer to Main than to Simple, but
the angles are large enough, especially when compared to the Simple/Main
column, to discourage any attempt at explaining the syntax of Main, or Simple,
based on the syntax of well-edited journalistic prose. We conclude that the
simplicity of Simple, evident both from reading the material and from the
Gunning Fog index discussed above, is due primarily to Main having
considerably longer sentences. A secondary effect may be the use of shorter
subsentences (comma-separated stretches) as well, but this remains unclear in
that the number of subsentence separators (commas, colons, semicolons, parens,
quotation marks) per sentence is considerably higher in Main (1.62) than in
Simple (1.01), so a Main subsentence is on the average not much longer than a
Simple subsentence (8.62 vs 7.96 content words/subsentence).

\begin{table}[ht!]
\caption{
\bf{Statistical similarity between different samples at different length of n-grams.}}
\begin{tabular}{cccc}
n & SM & MW & SW\\
2 & 13.1 & 28.3 & 33.8 \\
3 & 16.5 & 33.4 & 40.4 \\
4 & 20.1 & 40.8 & 49.8 \\
5 & 28.7 & 47.9 & 58.2 \\
\end{tabular}
\begin{flushleft}
Angle, in decimal
degrees, between Simple and Main (column SM), Main and WSJ (column MW), and
Simple and WSJ (column SW) based on POS n-grams for $n= 2,\ldots,5$, under
condition SN, with postprocessing of n-grams spanning sentence boundaries.
\end{flushleft}
\label{tab:angle}
\end{table}

\subsection*{Topical comparison}

Clearly, readability of text is a very context dependent feature. The more
conceptually complex a topic, the more complex linguistic structures and
the less readability are expected. To examine this intuitive hypothesis, we
considered different articles in different topical categories.  Instead of
systematically covering all possible categories of articles, here we
illustrate the phenomenon on a limited number of cases, where significant
differences are observed. The readability index of 10 selected articles from different topical categories is measured and reported in
in Table~\ref{tab:example}.

\begin{table}[ht!]
\caption{
\bf{Comparison of readability in Main and Simple English Wikipedias}}
\begin{tabular}{ccc}
Article   & $F_{{\rm Main}}$    &  $F_{{\rm Simple}}$ \\
Philosophy    &    16.6    &    11.3    \\
Physics    &    15.9    &    11.1    \\
Politics    &    14.1    &    8.9    \\
You're My Heart, You're My Soul    (song)&    9.6    &    5.8    \\
Real Madrid C.F.    &    11.6    &    7.6    \\
Immanuel Kant    &    15.7    &    10.3    \\
Albert Einstein    &    13.5    &    8.9    \\
Barack Obama    &    12.7    &    9.7    \\
Madonna (entertainer)    &    11.2    &    8.9    \\
Lionel Messi    &    12.8    &    7.9    \\
\end{tabular}
\begin{flushleft}Gunning fog index for the same example articles in Main and Simple.
\end{flushleft}
\label{tab:example}
\end{table}

While these results are clearly indicative of the main tendencies, for more
reliable statistics we need larger samples. To this end we sampled
over $\sim 50$ articles from 10 different categories and averaged the
readability index for the articles within the category. Results are shown in
Table~\ref{tab:topic}.  The numbers make it clear that more sophisticated
topics, e.g. {\it Philosophy} and {\it Physics} require more elaborate
language compared to the more common topics of {\it Politics} and {\it Sport}.
In addition, there is considerable difference between subjective and objective
articles, in that the level of complexity is slightly higher in the former:
more objective articles (e.g. biographies) are more readable.

\begin{table}[ht!]
\caption{
\bf{Readability in different topical categories}}
\begin{tabular}{ccc}
  Category   & $F_{{\rm Main}}$    &  $F_{{\rm Simple}}$ \\
Philosophy & 17.2$\pm0.6$ & 12.7$\pm0.8$ \\
Physics & 16.5$\pm0.4$ & 11.3$\pm0.7$  \\
Politics & 14.0$\pm0.5$ & 11.2$\pm0.8$ \\
Songs & 13.3$\pm0.6$ & 11.0$\pm0.7$  \\
Sport clubs &  12.2$\pm0.7$ & 10.1$\pm0.6$ \\
Philosophers & 15.9$\pm0.6$& 11.5$\pm0.8$  \\
Physicists & 15.0 $\pm0.5$ & 10.0$\pm0.7$ \\
Politicians & 13.1$\pm0.4$ & 10.2$\pm0.6$\\
Singers &  13.2$\pm0.4$ & 10.1$\pm0.5$  \\
Athletes & 13.1$\pm0.3$ & 10.1$\pm0.6$ \\
 \end{tabular}
\begin{flushleft}Gunning fog index for samples of articles in 10 different categories in Main and Simple.
\end{flushleft}
\label{tab:topic}
\end{table}

\subsection*{Conflict and controversy}

Wikipedia pages usually evolve in a smooth, constructive manner, but sometimes
severe conflicts, so called {\it edit wars}, emerge. A measure $M$ of
controversially was coined by appropriately weighting the number of mutual
reverts with the number of edits of the participants of the conflict in our
previous works \cite{sumi2011a,sumi2011b,yasseri2012}.  (For the exact
definition and more details, see Text S1 in the Supporting Information.) By
measuring $M$ for articles, one could rank them according to controversiality
(the intensity of editorial wars on the article).

In order to enhance the collaboration, resolve the issues, and discuss the
quality of the articles, editors communicate to each other through the ``talk
pages'' \cite{wiki-talk} both in controversial and in peacefully evolving
articles.  Depending on the controversially of the topic, the language that is
used by editors for these communications can become rather offensive and
destructive.  

In classical cognitive sociology \cite{deutsch1973}, there is a distinction
between ``constructive'' and ``destructive'' conflicts.  ``Destructive
processes form a coherent system aimed at inflicting psychological, material
or physical damage on the opponent, while constructive processes form a
coherent system aimed at achieving one's goals while maintaining or enhancing
relations with the opponent'' \cite{samson2010}. There are many
characteristics that distinguish these two types of interactions, such as the
use of swearwords and taboo expressions, but for our purposes the most
important is the lowering of language complexity in the case of destructive
conflict \cite{samson2010}.

Since we can locate destructive conflicts in Wikipedia based on measuring $M$,
a computation that does not take linguistic factors into account, we can check
independently whether linguistic complexity is indeed decreased as the
destructivity of the conflict increases. To this end, we created two similarly
sized samples, one composed of 20 highly controversial articles like {\it
  Anarchism} and {\it Jesus}, the other composed of 20 peacefully developing
articles like {\it Deer} and {\it York}.  The Gunning fog index was calculated
both for the articles and the corresponding talk pages for both
samples. Results are shown in Table~\ref{tab:conflict}.  We see that the fog
index of the conflict pages is significantly higher than those of the peaceful
ones (with 99.9\% confidence calculated with Welch's t-test). This is in
accord with the previous conclusion about the topical origin of differences in
the index (see Table~\ref{tab:topic}): clearly, conflict pages are usually
about rather complex issues. 

\begin{table}[ht!]
\caption{
\bf{Controversy and readability}}
\begin{tabular}{ccc}
    & Controversial &    Peaceful \\

$F_{{\rm Article}}$ & 16.5$\pm0.9$ & 11.6$\pm0.4$ \\
$F_{{\rm Talk}}$ & 11.7$\pm0.6$ & 8.6$\pm0.8$ \\
$\Delta F=F_{{\rm Article}}-F_{{\rm Talk}}$ & 4.8& 3.0\\
\end{tabular}
\begin{flushleft}Gunning fog index for two sample articles of highly controversial and peaceful articles and the corresponding talk pages.
\end{flushleft}
\label{tab:conflict}
\end{table}

In both samples there is a notable decrease in the fog index when going from
the main page to the talk page, but this decrease is considerably larger for
the conflict pages (4.8 vs. 3.0, separated within a confidence interval of
85\%).  This is just as expected from earlier observations of linguistic
behavior during destructive conflict \cite{samson2010}. The language
complexities for controversial articles and the corresponding talk pages are
higher to begin with, but the amount of reduction in language complexity
$\Delta F$ is much more noticeable in the presence of destructive conflicts
and severe editorial wars. 

\subsection*{Conclusions and future work}

In this work we exploited the unique near-parallelism that obtains between the
Main and the Simple English Wikipedias to study empirically the linguistic
differences triggered by a single stylistic factor, the effort of the editors
to make Simple simple. We have found, quite contrary to naive expectations,
and to Simple Wikipedia guidelines, that classic measures of vocabulary
richness and syntactic complexity are barely affected by the simplification
effort. The real impact of this effort is seen in the less frequent use of
more complex words, and in the use of shorter sentences, both directly
contributing to a decreased Fog index.

Simplification of the lexicon, as measured by $C$ or word entropy, is hardly
detectable, unless we directly compare the corresponding Simple and Main
articles, and even there the effect is small, 3.4\%. The amount of syntactic
variety, as measured by POS n-gram entropy, is decreased from Main to Simple
by a more detectable, but still rather small amount, 2-3\%, with an estimated
20-30\% of this decrease due to robotic generation of pages. Altogether, the
complexity of Simple remains quite close to that of newspaper text, and very
far from the easily detectable simplification seen in spoken language.

We believe our work can help future editors of the simple Wikipedia, e.g.  by
adding robotic complexity checkers. Further investigation of the linguistic
properties of Wikipedias in general and the simple English edition in
particular could provide results of great practical utility not only in
natural language processing and applied linguistics, but also in foreign
language education and improvement of teaching methods. The methods used here
may also find an application in the study of other purportedly simpler
language varieties such as creoles and child-direceted speech. 

\section*{Acknowledgments}
TY thanks Katarzyna Samson for useful discussions. We thank Attila
  Zs\'{e}der and G\'{a}bor Recski for helping us with the POS analysis.
Suggestions by the anonymous PLoS ONE referees led to significant 
improvements in the paper, and are gratefully acknowledged here. 

\section*{Supporting Information}

\subsection*{Text S1: Controversy measure}

To quantify the controversiality of an article based on its editorial history,
we focus on ``reverts'', i.e. when an editor undoes another editor's edit
completely. To detect reverts, we first assign a MD5 hash code \cite{MD5} to
each revision of the article and then by comparing the hash codes, detect when
two versions in the history line are exactly the same. In this case, the
latest edit (leading to the second identical revision) is marked as a revert,
and a pair of editors, namely a reverting and a reverted one, are
recognized. A ``mutual revert'' is recognized if a pair of editors $(x,y)$ is
observed once with $x$ and once with $y$ as the reverter. The weight of an editor
$x$ is defined as the number of edits $N$ performed by her, and the weight of a
mutually reverting pair is defined as the minimum of the weights of the two
editors. The controversiality $M$ of an article is defined by summing the 
weights of all mutually reverting editor pairs, excluding the topmost pair, 
and multiplying this number by the total number of editors $E$ involved in 
the article. In formula,

\begin{equation}
M = E\sum_{\text{all mutual reverts}} min(N^{\rm d}, N^{\rm r}),
\end{equation}

where $N^{\rm r/d}$ is the number of edits on the article committed by
reverting/reverted editor. The sum is taken over mutual reverts rather than
single reverts because reverting is very much part of the normal workflow,
especially for defending articles from vandalism. The minimum of the two
weights is used because conflicts between two senior editors contributing more
to controversiality than conflicts between a junior and a senior editor, or
between two junior editors. For more details on how the above formula defining
$M$ was selected and validated see \cite{yasseri2012} and especially {\it Text
  S1} in its Supporting Information.

\subsection*{Text S2: Corpora and analysis tools}

To download Wikipedia dumps use the static snapshots from
\url{http://dumps.wikimedia.org}. To download the dynamic content, especially
the most updated version of individual articles, use the ``MediaWiki API''
online platform accessible at
\url{http://www.mediawiki.org/wiki/API:Main\_page}. The Brown, Switchboard,
and WSJ corpora are distributed by the Linguistic Data Consortium as part of
the Penn Treebank,
\url{http://www.ldc.upenn.edu/Catalog/catalogEntry.jsp?catalogId=LDC99T42} The
POS tagging of these texts, while not necessarily 100\% correct, is manually
corrected and generally considered a gold standard against which POS taggers
are evaluated. Many gigaword corpora (including Arabic, Chinese, English,
French, and Spanish) are available from the LDC, see
\url{http://www.ldc.upenn.edu/Catalog/catalogSearch.jsp} 

To clean the text from Wikimedia tags and external references, we used the
WikiExtractor developed at the University of Pisa Multimedia Lab, available at
\url{http://medialab.di.unipi.it/wiki/Wikipedia_Extractor}.  Another system
with similar capabilities is ``wiki2text''
\url{http://wiki2text.sourceforge.net}. We used faster (flex-based) versions
of the original Koehn tokenizer and Mikheev sentence splitter, available at
\url{https://github.com/zseder/webcorpus}.

For English stemming, the standard is the ``Porter Stemming Algorithm''
\url{http://tartarus.org/~martin/PorterStemmer}. For other languages a good
starting point is
\url{http://aclweb.org/aclwiki/index.php?title=List_of_resources_by_language}.

We calculated the Gunning Fog index using the code and algorithm of Greg Fast
\url{http://cpansearch.perl.org/src/GREGFAST/Lingua-EN-Syllable-0.251/Syllable.pm}.
For part-of-speech tagging we used the ``HunPOS tagger''
\url{http://code.google.com/p/hunpos/} and the ``HunNER NE recognizer'', which
are specific applications of the ``HunTag tool'', available at
\url{https://github.com/recski/HunTag/}.

To perform the n-gram analysis we used the ``N-Gram Extraction Tools''
\url{http://homepages.inf.ed.ac.uk/lzhang10/ngram.html} of Le Zhang.

All the abovementioned code and packages are open source and available
publicly under GPL, LGPL, or similar licenses, but some corpora may have
copyright restrictions. 

\bibliography{simple}

\end{document}